\pdfoutput=1

\documentclass[11pt]{article}

\usepackage[]{acl}

\usepackage{times}
\usepackage{latexsym}
\usepackage{amsmath}

\usepackage[T1]{fontenc}

\usepackage[utf8]{inputenc}

\usepackage{microtype}
\usepackage{multirow}
\usepackage{graphicx}
\usepackage{makecell}

\newcommand{\relregmodel}{\textsc{RelReg}}
\newcommand{\multiencmodel}{\textsc{SegEnc}}
\newcommand{\led}{\textsc{LED}}
\newcommand{\marge}{\textsc{MaRGE}}
\newcommand{\bart}{\textsc{BART}}
\newcommand{\dpr}{\textsc{DPR}}
\newcommand{\twotower}{\textsc{RelRegTT}}
\newcommand{\dyle}{\textsc{DYLE}}
\newcommand{\summn}{\textsc{SUMM\textsuperscript{N}}}
\newcommand{\lead}{\textsc{Lead}}
\newcommand{\goldspans}{\textsc{Gold Spans}}
\newcommand{\wikisuffix}{\textsc{-W}}

\title{Exploring Neural Models for Query-Focused Summarization}
\newcommand*{\affaddr}[1]{#1}

\author{
        \textbf{Jesse Vig}\Thanks{~~Equal contribution}
  \quad \textbf{Alexander R. Fabbri}\textsuperscript{$*$}
  \quad \textbf{Wojciech Kry\'sci\'nski}\textsuperscript{$*$}\\
  \textbf{Chien-Sheng Wu}
  \quad \textbf{Wenhao Liu}\\
  \affaddr{Salesforce Research} \\
  \{jvig, afabbri, kryscinski, wu.jason, wenhao.liu\}@salesforce.com
}

\begin{document}
\maketitle
\begin{abstract}
Query-focused summarization (QFS) aims to produce summaries that answer particular questions of interest, enabling greater user control and personalization.
While recently released datasets, such as QMSum or AQuaMuSe, facilitate research efforts in QFS, the field lacks a comprehensive study of the broad space of applicable modeling methods. 
In this paper we conduct a systematic exploration of neural approaches to QFS, considering two general classes of methods: two-stage extractive-abstractive solutions and end-to-end models.
Within those categories, we investigate existing models and explore strategies for transfer learning.
We also present two modeling extensions that achieve state-of-the-art performance on the QMSum dataset, up to a margin of 3.38 ROUGE-1, 3.72 ROUGE-2, and 3.28 ROUGE-L when combined with transfer learning strategies. 
Results from human evaluation suggest that the best models produce more comprehensive and factually-consistent summaries compared to a baseline model.
Code and checkpoints are made publicly available: \url{https://github.com/salesforce/query-focused-sum}.
\end{abstract}
\section{Introduction}\label{sec:introduction}
Text summarization aims at transforming long documents into short snippets that contain only the most important information from the source document.
The field has seen substantial progress driven by the availability of large-scale models pre-trained on vast amounts of data~\citep{devlin2018bert, lewis2019bart}, the development of summarization-specific pre-training strategies~\citep{zhang2019pegasus, zhao_seal_2020}, and computationally efficient neural architectures~\citep{zaheer2020bigbird}.

The majority of recent research efforts in text summarization assume an unconstrained setting in which models are given only a source document as input and are expected to generate a general summary covering the salient aspects from the source.
The performance of such models has been evaluated on benchmark datasets spanning various domains: news articles~\citep{nallapati2016cnndm, narayan2018xsum, fabbri2019multinews}, legal documents~\citep{sharma2019bigpatent}, scientific writing~\citep{cohan2018arxivpubmed}, or creative writing~\citep{kryscinski2021booksum, chen2021summscreen}.
However, it has been shown that summarization in an unconstrained setting is an ill-defined task where multiple generated summaries are equally relevant~\citep{kryscinski2019critical}.
This in turn hinders the ability to evaluate and understand the models' content selection capacity.
In addition, such generic summarization models lack control mechanisms that would allow end users to customize summaries to their particular needs and expectations.

\textit{Query-focused summarization} (QFS) is a subtask within text summarization that focuses on generating summaries where the summary content is tailored to a user-specified query that is passed alongside the source document as input to the model.
Each source document can be associated with multiple unique queries inquiring about different information from that document.
In this setting, end users are enabled to explicitly specify their preferences for the summary, and the relevance of the output summary may be evaluated more precisely with respect to the input query.
Research on this task has been accelerated by the recently introduced high-quality datasets, such as QMSum~\citep{zhong-etal-2021-qmsum} and AQuaMuSe~\citep{kulkarni2020aquamuse}.

In this work we conduct a systematic, exploratory study of different approaches to query-focused text summarization, considering both two-step and end-to-end neural methods. 
We present two models, \relregmodel~and \multiencmodel, which achieve state-of-the-art ROUGE scores on the QMSum dataset, up to a margin of 3.38 R-1, 3.72 R-2, and 3.28 R-L when combined with transfer learning methods.
The \relregmodel~model uses a two-step approach to solving the problem, where the first step extracts content relevant to the given query and the next step synthesizes the extracted fragments into a coherent summary.
The \multiencmodel~method follows an end-to-end framework in which individual document segments are separately encoded to avoid the computational bottleneck of long input documents, and the decoder jointly attends to all encoded segments when producing the summary.
Through quantitative studies, we compare our models with other baselines and discuss the trade-offs of the end-to-end methods and pipelined approaches.
We also perform human evaluation to understand the qualitative differences between the models.
Together with this manuscript, we share the code base and model checkpoints to enable future research in this area.

\section{Related Work}\label{sec:related-work}

\subsection{Query-Focused Summarization}\label{ssec:related-qfs}
Query-focused summarization aims to generate a summary of a given text conditioned upon a query. 
Initial work in this area centered around unsupervised extractive approaches \cite{wan2007manifold,litvak-vanetik-2017-query} due to the limited availability of task-specific training data \cite{dang2005overview}. 
More recent work has taken advantage of the relationship between query-focused summarization and the more data-rich task of question answering for extractive summarization \cite{egonmwan-etal-2019-cross}, reranking documents within a retrieval pipeline \cite{su-etal-2020-caire}, and abstractive summarization \cite{su-etal-2021-improve,baumel_query_2018,xie2020conditional}.
\citet{xu-lapata-2020-coarse} introduce a pipeline consisting of a relevance estimator filter followed by query-focused evidence and centrality estimators, while other work converts generic summarization dataset to query-focused training data \cite{xu-lapata-2021-generating} or performs latent query modeling \cite{xu_latent_queries_2021}.

Recently, several query-focused summarization datasets have been introduced, which can be further divided into short-document datasets, whose source document length does not exceed the input limits of standard pretrained models, and long-document datasets.
Within short-document, query-focused summarization, AnswerSumm \cite{fabbri2021answersumm} is composed of summaries of answers to queries from StackExchange forums, while WikiHowQA \cite{DengLXCL0S20} proposes the task of answer selection followed by the summarization of individual response articles to queries from the how-to site WikiHow.
Within long-document summarization, WikiSum \cite{liu2018generating} consists of Wikipedia article titles as queries, the first paragraph of the article as the summary, and documents referenced by the article as the input.
AQuaMuSe \citep{kulkarni2020aquamuse} is a query-focused multi-document summarization dataset with user-written queries and human-verified long-answer summaries from the Natural Questions dataset~\citep{kwiatkowski2019naturalquestions}, and QMSum \cite{zhong-etal-2021-qmsum} is a manually-curated dataset for query-focused dialog summarization.
QMSum and AQuaMuSe are of particular interest to our study due to the combined challenges of query-focused and long-document summarization and the presence of high-quality, curated query-summary pairs. 

Recent work on QMSum has introduced task-specific denoising objectives for meeting summarization \cite{zhong2021dialoglm}, generated final fine-grained summaries based on multiple coarse-grained steps \cite{zhang2021summn}, and treated the extractive text of an extractive-abstractive model as a latent variable \cite{mao2021dyle}.
\citet{zhang2021exploratory} analyze the challenges of long dialogue summarization such as the input length, the role of queries, and domain adaptation.
Our work builds on QA-motivated methods and presents two approaches yet to be applied in query-focused summarization that each achieve state-of-the-art results, including a two-step model and an end-to-end model.

\subsection{Long Document Summarization}
Long document summarization addresses the setting where source document length exceeds the input limits of standard pre-trained models. 
Approaches to this task can largely be divided into two categories: two-step extractive-abstractive frameworks, which first extract a subset of the text as input to an abstractive model, and end-to-end models, which process the input within a single model.
The two-step pipeline has been applied to topic-focused Wikipedia summarization \cite{liu_wikisum_2018,liu-lapata-2019-hierarchical,perez-beltrachini-etal-2019-generating}, low-resource summarization \cite{bajaj-etal-2021-long}, and single-document summarization \citet{chen-bansal-2018-fast}.
End-to-end approaches address the input-length problem using sparse-attention models. \citet{beltagy2020longformer} introduce the Longformer, consisting of local attention as well as global attention between select input tokens. Other approaches make use of dynamic attention mechanisms \cite{zhao_seal_2020,manakul2021sparsity,cui-hu-2021-sliding}, sliding window strategies \cite{liu2021dynamic}, and other mechanisms to introduce sparsity into the model \cite{huang-etal-2021-efficient,liu2021hetformer}.
\citet{izacard-grave-2021-leveraging} concatenate the outputs of multiple encoders as input to a generator component for the task of open domain question answering.
In our work we build on these models for query-focused summarization and perform extensive hyperparameter ablations, achieving state-of-the-art results over other two-step and end-to-end models.

\section{Methodology}\label{sec:methodology}
We present existing methods and propose modeling extensions to address the challenges of query-focused summarization.

\subsection{Two-Step Approaches}
Two-step approaches consist of an \textit{extractor} model, which extracts parts of the source document relevant to the input query, and an \textit{abstractor} model, which synthesizes the extracted segments into a final summary.
We consider \textit{score-and-rank} extractor models, which first score each source passage for relevance to the query and then rank the passages in descending order of relevance, with the concatenated and truncated results passed to the abstractor.
In this work we present two types of scoring models: \textit{single-encoder} models and \textit{dual-encoder} models, which we describe below. 
All two-step approaches share the same abstractor, a BART-large model.

\subsubsection{Single-Encoder Models}
Single encoder models concatenate a query and source passage as input to the scoring function that produces the similarity score.
Those models benefit from full cross-attention between query and passage, resulting in richer data representations.
\paragraph{\marge} \cite{xu-lapata-2021-generating}  is a single-encoder, \textbf{Ma}sked \textbf{R}OU\textbf{GE} extractor that aims to improve upon low-resource query-focused summarization by synthesizing query-focused data from more resource rich, generic summarization datasets. 
This model is trained to predict the relevance of each passage in the source document with respect to a query, where the proxy for relevance is the ROUGE overlap between the passage and the reference summary.
For training on generic summarization datasets, \marge~ uses pseudo-queries that are created by masking content words in the  reference summaries. 

When performing inference using real queries, certain query words (e.g., wh-words) are masked to better align the queries to the pseudo-queries from the training process.
Following \citet{xu-lapata-2021-generating}, we apply \marge~trained for masked relevance prediction on Multi-News \cite{fabbri-etal-2019-multi} without training on our target dataset.
\paragraph{\relregmodel} 
Motivated by the retrieval component of \marge, we propose the \relregmodel~(RELevance REGression) model, which trains a relevance prediction model directly on QFS data using the \textit{original, non-masked} query. 
Like \marge, this model is trained to predict the ROUGE overlap between a source passage and the reference summary, using only the passage and query as input. 
A single-encoder model jointly encodes the delimiter-separated query and passage, and the final layer of the model outputs the predicted relevance value.

\subsubsection{Dual-Encoder Models}
Dual-encoder models separately encode a query and source passage before calculating the cosine similarity between the embeddings to compute the relevance score.
This class of models offers computational benefits, as passage embeddings may be precomputed and stored for a given input, while the single-encoder model must be run over all passages should a new query be introduced.

\paragraph{\dpr}~\cite{karpukhin-etal-2020-dense} is a dual-encoder model that separately encodes queries and passages into an embedding space optimized for calculating semantic similarity between the two, showing improved results over traditional vector-space models. 
We fine-tune a \dpr~extractor model directly on the target dataset. 
As opposed to other locators that optimize with respect to the continuous ROUGE overlap, \dpr~uses the ROUGE score between the passage and reference summary to identify binary positive and negative passages and optimizes the negative log likelihood of the positive passages. 

\paragraph{\twotower} (RELevance REGression Two Tower) is a more computationally-efficient version of \relregmodel~ that uses a dual-encoder architecture to predict ROUGE-based relevance scores.
This model is implemented with a backbone architecture of Sentence-BERT \cite{reimers-gurevych-2019-sentence}, using a shared-parameter encoder for each of the query and passage and a special token appended to each input that identifies it is as either query or passage, following the suggested best practices of \citet{reimers-gurevych-2019-sentence}. 
The final output for the model is based on the inner product of the pooled embeddings for the query and passage.

\subsection{End-to-End Approaches}\label{ssec:end-to-end-methods}
Two-step pipelines depend on the strength of the retrieval component, and may still fail to capture all relevant content despite an ideal retriever, due to length limitations of the generation component. This motivates our experiments on end-to-end models that can incorporate longer input texts. 

\paragraph{\bart}\citep{lewis2019bart}
As a baseline end-to-end model, we consider \bart, an encoder-decoder Transformer model pre-trained using a denoising objective.
\bart~is composed of a bidirectional encoder module and an autoregressive decoder model that attends to the encoder's final layer outputs.
Due to the quadratic memory complexity of the encoder's full self-attention mechanism, the model input size is limited to 1024 tokens.
In our experiments, we prepare the input to \bart~ by concatenating the query, a delimiter token, and the source document, and then truncating the combined text to the model's input size.

\paragraph{\led}
To circumvent the input size limitations of the \bart~ model, we include the Longformer Encoder-Decoder~\cite{beltagy2020longformer} (\led) in our study
\led~replaces the quadratic self-attention mechanism of traditional Transformers with a memory-efficient version that combines local attention with sparse global attention.
The architecture allowed us to run experiments with input sizes up to 16384 tokens.
Based on insights from the original work on tuning the model to the QA task, we configure the global attention mechanism to span the entire query.

\paragraph{\multiencmodel}
We also consider a simpler form of sparse attention in the encoder based solely on windowed local attention, combining elements of \led~with  Fusion-in-Decoder (FiD) \cite{izacard-grave-2021-leveraging}, a model for open-domain question answering.
In our Segment Encoder (\multiencmodel) model, the source document is split into fixed-length overlapping\footnote{We use segments that are 50\% overlapping, though other configurations may be considered.} segments, each of which is separately appended to the query and encoded using a standard Transformer model. 
Similar to FiD, these encodings are then concatenated into a single embedding sequence and passed to a decoder model that generates the summary.
Since there is no cross-attention between the encoded segments, the attention mechanism scales linearly in the number of segments and hence the length of the source document.
Nonetheless, the decoder can attend to all encoded segments jointly, enabling the encoder-decoder architecture to operate in an end-to-end fashion.  
This model is motivated by two hypotheses: 1) query-relevant sections within a source document are often small enough to be processed by standard Transformer models (e.g. 1024 tokens), and 2) each query-relevant section may be understood independently of other sections, removing the need for cross-attention between the segments.

\subsection{Data}
We analyze our methods on two high-quality query-focused, long-document datasets. 
\par \noindent
\label{ssec:data}
\textbf{QMSum}~\cite{zhong-etal-2021-qmsum} is a query-focused dialogue summarization dataset consisting of 1,808 query-summary pairs over 232 meetings from product design, academic, and political committee meetings, all conducted in English.
QMSum also includes additional annotations such as topic segmentations and highlighted text spans associated with reference summaries.
We leverage the provided span annotations to run oracle experiments.
We focus our analysis on QMSum due to the availability of prior work as points of comparison. 
\par
\noindent \textbf{AQuaMuSe}~\cite{kulkarni2020aquamuse} is a query-focused multi-document summarization dataset consisting of 5,519 query-long answer summary pairs from the Natural Questions question-answering dataset \cite{kwiatkowski2019naturalquestions} and associated input documents from the Common Crawl\footnote{\url{https://commoncrawl.org/}}.
Input documents for the original dataset were selected based on embedding similarity with respect to the summary, and hyperparameters can be chosen to control the level of semantic overlap between the input document set and the summary. 
Data replication details are found in the Appendix. 
We use AQuaMuSe to examine the generalizability of our QMSum results.

\subsection{Experiment Setup}
\begin{table*}[t!]
\centering
\resizebox{\textwidth}{!}{%
\begin{tabular}{l|cccc|cccc|cccc|cccc||cc}
\hline
& \multicolumn{16}{c||}{\makecell{\textbf{Lexical Overlap b/w} \\ \textbf{Extractors and References}}} & \multicolumn{2}{c}{\makecell{\textbf{Span Overlap b/w} \\ \textbf{Extractors and Golden Spans}}} \\ 
 \cline{2-19}
\textbf{Model} & \multicolumn{4}{c|}{\textbf{Top-1}} & \multicolumn{4}{c|}{\textbf{Top-5}} & \multicolumn{4}{c|}{\textbf{Top-15}} & \multicolumn{4}{c||}{\textbf{All}} & \multicolumn{2}{c}{\textbf{All}} \\
 & \textit{R-1} & \textit{R-2} & \textit{R-L} & $\bar{x}$ & \textit{R-1} & \textit{R-2} & \textit{R-L} & $\bar{x}$ & \textit{R-1} & \textit{R-2} & \textit{R-L} & $\bar{x}$ & \textit{R-1} & \textit{R-2} & \textit{R-L} & $\bar{x}$ & \textit{Precision} & \textit{Recall} \\ \hline
\goldspans & 15.00 & 3.80 & 11.10 & 60 & \textbf{20.89} & \textbf{6.05} & \textbf{15.04} & 218 & \textbf{19.62} & \textbf{5.99} & \textbf{14.28} & 386 & \textbf{16.09} & \textbf{5.60} & \textbf{12.47} & 660 & 0.75 & 1.00
 \\
\lead & 8.17 & 0.98 & 6.30 & 82 & 12.84 & 1.69 & 9.17 & 309 & 13.13 & 1.81 & 9.21 & 463 & 8.77 & 1.79 & 6.77 & 978 & 0.09 & 0.20 \\ \hline
\dpr & 11.31 & 1.99 & 8.72 & 34 & 17.46 & 2.86 & 12.21 & 156 & 15.38 & 2.74 & 10.64 & 394 & 9.75 & 2.23 & 7.42 & 932 & 0.22 & 0.27 \\
\twotower & 23.67 & 3.34 & 15.66 & 82 & 16.13 & 3.35 & 11.18 & 413 & 9.65 & 2.58 & 7.31 & 930 & 9.16 & 2.52 & 6.99 & 994 & 0.07 & 0.24 \\ \hline
\marge & 7.13 & 0.72 & 5.81 & 20 & 13.76 & 1.39 & 10.22 & 92 & 14.85 & 1.74 & 11.09 & 269 & 9.21 & 1.52 & 7.16 & 896 & 0.15 & 0.21 \\
\relregmodel & \textbf{24.57} & \textbf{4.33} & \textbf{16.57} & 88 & 17.52 & 4.11 & 12.21 & 418 & 10.56 & 3.04 & 8.06 & 884 & 9.62 & 2.87 & 7.47 & 989 & 0.11 & 0.28 \\ 
\hline
\end{tabular}
}
\caption{
Performance of extractor models on the QMSum validation set.
The left section presents the lexical overlap between the utterances retrieved by extractor models and the reference summaries, evaluated by means of ROUGE-1 (\textit{R-1)}, ROUGE-2 (\textit{R-2}), and ROUGE-L (\textit{R-L}) metrics.
Segments of the section focus on the lexical overlap between the highest ranked 1 (Top-1), 5 (Top-5), 15 (Top-15) utterances, and all utterances truncated to a 1024 token limit (All).
The table also includes the average word counts of all extracted utterances, denoted as $\bar{x}$.
The right section shows the span overlap between the utterance spans retrieved by the extractor models and those collected from human annotators by the authors of QMSum.
The performance is evaluated by means of \textit{Precision} and \textit{Recall} scores and uses the highest ranked utterances truncated to the limit of 1024 tokens.
}
\label{tab:extractor_models}
\end{table*}

\paragraph{Implementation}
Models were implemented using the PyTorch~\citep{paszke2020pytorch} and Huggingface~\citep{wolf2019huggingface} libraries.
Model weights were initialized from pre-trained checkpoints available through the Huggingface Model Hub\footnote{\url{https://huggingface.co/models}}.
All \bart~models were based on the \texttt{facebook/bart-large} checkpoint, the LED-model was based on the \texttt{allenai/led-large-16384}~checkpoint, which itself is based on \texttt{BART-large}.

\paragraph{Training \& Inference}
Models were trained for 10 epochs with final checkpoints selected based on the average of ROUGE-$\{1, 2, L\}$ (\textit{R-1}, \textit{R-2}, \textit{R-L}) scores achieved on the validation set.
Gradient checkpointing \cite{chen2016training} was used for the \led~and \multiencmodel~models to reduce the memory footprint.
Model outputs were decoded using beam search with 4 beams.
To ensure high consistency of results, all experiments in \S\ref{sec:results} were repeated 5 times with results averaged across runs.

\paragraph{Evaluation}
Models were automatically evaluated using the ROUGE-$\{1, 2, L\}$ metrics~\citep{lin2004rouge} included in the SummEval toolkit~\citep{fabbri2021summeval}.
Models were also manually evaluated by hired human annotators.
Annotators were hired through the Amazon Mechanical Turk platform.
Workers were selected from English speaking countries and offered an hourly rate of approximately 12 USD.
The study was conducted on 50 model generated examples chosen at random from the test set of QMSum.

\section{Model Exploration}\label{sec:results}
In this section, we first analyze the effects of model-specific architectural and hyperparameter choices on the performance of two-stage (\S\ref{ssec:two-stage-results}) and end-to-end models (\S\ref{ssec:end-to-end-results}). Next, we study the task-specific knowledge transfer capabilities of different pre-training strategies in \S\ref{ssec:domain-adaptation}. Lastly, we conduct a final evaluation and comparison of all discussed models in \S\ref{ssec:final-results}. 
All experiments and analyses presented in this section were conducted on QMSum.
\subsection{Two-Stage Approaches}\label{ssec:two-stage-results}
For two-stage models, we first focus on evaluating the extractor component and comparing performance to baseline heuristics.
We quantify extractor performance using two metrics: 1) lexical overlap between the extracted utterances and reference summaries, computed using R-1, R-2, and R-L metrics, 2) span overlap between the extracted and golden spans included with QMSum represented by Precision and Recall scores, with results shown in Table~\ref{tab:extractor_models}.
In both cases, we first order utterances of the conversation according to the scores assigned by the extractor models, then concatenate the utterances and finally truncate the result to 1024 tokens (excluding the space reserved for the query) to mimic the input length limits of downstream abstractor models; we present those numbers as the \textit{All} columns in the table.
For the lexical overlap, we also show the scores for the best 1 (\textit{Top-1}), 5 (\textit{Top-5}), and 15 (\textit{Top-15}) utterances.

The results show that the best-performing extractor  model is \relregmodel~closely followed by \twotower~in the \textit{Top-1} evaluation and DPR in the \textit{Top-5}, \textit{Top-15}, and \textit{All} cases.
We note that both the \relregmodel~and \twotower~models tend to select longer utterances than the other extractors; the regression-based training mirrors the ROUGE overlap score which favors longer, more informative utterances.
However, despite their strong performance in extracting top-matching utterances, the results also expose a considerable gap between model-based approaches and human annotations when considering the entirety of extracted spans.
This shows a promising topic for future work in this matter.
We also notice that despite the simplicity of the \lead~heuristic, which  extracts the first $k$ utterances in their original order, it remains competitive with the data-driven extractor models when we consider the \textit{All} case.
An extended version of this study, which includes the lexical overlap between extracted spans and input queries is presented in Table~\ref{tab:extractor_models_full} in the Appendix.

Next, we analyze how the performance of the extractor components carries over to the final summarization task.
For the best-performing model, we additionally test the effect of varying the input segment size used during training and inference between 256 and 512 tokens.
Validation-set results for all models are reported in Table \ref{tab:extractor_summ_val}. 

We find that \dpr~slightly outperforms \twotower~for dual-encoder models.
 Among single-encoder models,  \relregmodel~ outperforms \marge~ by over a full R-1 point, which may explained by \relregmodel~using more direct supervision based on an in-domain query, rather than creating synthetic queries from an external dataset using masking.
We find that the single-encoder \relregmodel~ outperforms the best dual-encoder model; the cross-attention term in the single-encoder \relregmodel~ model allows it to better attend to the query when determining relevance. 
Intuitively, the ordering of results corresponds to the span overlap recall with the gold spans; the ability of the extractor to select produce high-recall rankings directly affects abstractor performance.
We see that increasing the input segment length used in training and inference for \relregmodel~ improves at 256 tokens but decreases at 512 tokens, suggesting that a balance is found between including additional context for ranking versus enabling a greater number of shorter segments that may capture more diverse content from the source. 

\begin{table}[t!]
\centering
\small
\begin{tabular}{lccc}
\hline \textbf{Model} & \textbf{R-1} & \textbf{R-2} & \textbf{R-L} \\ \hline
\dpr~ & 32.79 & 9.82 & 28.91 \\ 
\twotower~ &  32.65 & 9.00 & 28.57 \\ \hline
\marge~ & 31.90 & 9.10 & 28.17  \\ 
\relregmodel~ & 33.43 & 9.77 & 29.40 \\ 
\relregmodel~(256) & \textbf{34.67} & \textbf{11.53} & \textbf{30.66} \\ 
\relregmodel~(512) & 32.22 & 10.29 & 29.49 \\  \hline
\end{tabular}
\caption{Performance of two-step models on the QMSum validation set, divided into dual-encoder and single-encoder extractors. Input segment lengths are indicated in parentheses, and otherwise the model operates on utterance-level input.}
\label{tab:extractor_summ_val}
\end{table}

\subsection{End-to-End Approaches}\label{ssec:end-to-end-results}
We explore hyperparameter choices for two end-to-end architectures described in \S\ref{ssec:end-to-end-methods}: the Longformer Encoder-Decoder (\led) and Segment Encoder (\multiencmodel).
For both models, we consider different choices for input size (4096, 8192, or 16384 tokens) and attention window size\footnote{For \multiencmodel, attention window size is equivalent to segment size.} (256, 512, or 1024 tokens).
For \multiencmodel, we also consider two different segmentation strategies: overlapping segments (50\% overlap) and disjoint segments.
Validation set results for both models and a baseline \bart~ model are reported in Table~\ref{tab:end2end}.

\begin{table}[t!]
\centering
\small
\begin{tabular}{l|l|l|ccc}
\hline \textbf{Model} & \textbf{Input} & \textbf{Attn} & \textbf{R-1} & \textbf{R-2} & \textbf{R-L} \\ \hline
\bart~&1024 & 1024 & 32.42 & 9.62 & 28.37 \\ \hline
~& & 256 & 31.55 & 8.89 & 27.62 \\
~&4096& 512 & 32.25 & 9.27 & 28.29 \\
~&& 1024 & 32.16 & 9.05 & 28.27 \\ \cline{2-6} 
~&& 256 & 31.79 & 8.97 & 27.75 \\
\led~&8192& 512 & 32.76 & 9.38 & 28.65 \\
~&& 1024 & 32.85 & 9.26 & 28.73 \\ \cline{2-6} 
~&& 256 & 31.94 & 9.16 & 27.73 \\
~&16384& 512 & 32.88 & 9.82 & 28.90 \\
~&& 1024 & 32.98 & 9.60 & 29.08 \\ 
\hline
&  & 256 & 35.35 & 10.37 & 30.91 \\
& 4096 & 512 & 35.25 & 10.36 & 30.85 \\
& & 1024 & 34.36 & 9.85 & 30.13 \\ \cline{2-6} 
&& 256 & 36.51 & 11.36 & 31.87 \\
\multiencmodel~&8192& 512 & 36.68 & 11.71 & 32.08 \\
&& 1024 & 35.48 & 10.97 & 31.21 \\ \cline{2-6} 
&& 256 & 37.21 & 12.14 & 32.67 \\ 
&16384& 512 & \textbf{37.47} & \textbf{12.47} & \textbf{32.95} \\
&& 1024 & 36.30 & 11.71 & 32.01 \\ \hline
\multiencmodel-D&16384&512 & 36.68 & 11.97 & 32.35 \\
\hline

\end{tabular}
\caption{Performance of end-to-end models on the QMSum validation set, across varying input and attention window sizes (in number of tokens). \multiencmodel-D is a variant of \multiencmodel~in which the segments are \textit{disjoint} rather than overlapping; this ablation was evaluated on the best-performing \multiencmodel~hyperparameters.}
\label{tab:end2end}
\end{table}

We notice that both the \led~and \multiencmodel~benefit from increasing the input size and perform best with the input limit set to 16,384 tokens.
The optimal attention window for \led~is 1024, while \multiencmodel~performs best with an attention window of 512 tokens.
For \multiencmodel, using overlapping segments improves performance compared to using disjoint segments, suggesting that the additional context provided by the former approach is helpful for locating relevant content.
The \multiencmodel~model achieves the highest performance out of the end-to-end architectures with ROUGE scores of 37.47 \textit{R-1}, 12.47 \textit{R-2}, and 32.95 \textit{R-L} on the validation set.

The results also highlight that while the \led~model matches or slightly outperforms the \bart~baseline for higher maximum input and window sizes, it performs substantially worse than \multiencmodel.
This observation is consistent with prior findings on the QMSum dataset~\citep{zhang2021exploratory}.
One possible explanation for the lower performance of \led~ relative to \multiencmodel~is that \led~must adapt its parameters for a global attention mechanism that is absent from the backbone BART encoder model, whereas \multiencmodel~relies solely on local self-attention that is aligned with the backbone model.
This may be particularly relevant to QMSum given its relatively small size.

Practitioners may wish to consider the computational cost and efficiency of various hyperparameter settings.
Computational complexity increases with both input length and attention window size (since attention grows quadratically in attention-window size).
Complexity is also greater with the overlapping segment strategy compared to the disjoint segment strategy for the \multiencmodel~model, due to the greater number of resulting segments that are passed through the encoder and decoder modules.
\begin{table}[t!]
\centering
\small
\begin{tabular}{lccc}
\hline \textbf{Model} & \textbf{R-1} & \textbf{R-2} & \textbf{R-L} \\ \hline
No Transfer & 32.42 & 9.62 & 28.37 \\ 
AnswerSumm & 34.36 & 9.64 & 30.22 \\ 
AQuaMuse & 34.57 & 9.78 & 30.42 \\ 
WikiHowQA & 33.08 & 9.03 & 28.48 \\ 
CNNDM & 33.87 & 9.36 & 28.48 \\ 
WikiSum & \textbf{34.73} & \textbf{9.80} & \textbf{30.54} \\ \hline
\end{tabular}
\caption{QMSum validation-set performance of the end-to-end \bart~models first fine-tuned on related summarization tasks and then further fine-tuned on QMSum data. The model indicates the task first fine-tuned on, and input is truncated to 1024 tokens.}
\label{tab:task_transfer}
\end{table}

\subsection{Task-Specific Transfer}\label{ssec:domain-adaptation}
Having determined the best-performing models, we examine whether performance can be further improved by fine-tuning a model that has already been fine-tuned for a different summarization task. 
We conduct this study using the end-to-end \bart~on 1024 tokens, as this model is the backbone, albeit in varying ways, of both our two-step and end-to-end models. 
We test the transferring capabilities of models trained on the news summarization task from CNN/DailyMail \cite{nallapati2016cnndm}, which performed best among non query-focused datasets in \citet{zhang2021exploratory}. 
We also explore transferring from the previously-mentioned query- and topic-focused summarization tasks: AnswerSumm, AQuaMuSe, WikiHowQA, and WikiSum. 
We compare to fine-tuning from the original \bart~ checkpoint, with results shown in Table \ref{tab:task_transfer}. 

We find that transferring from any of the tasks improves over no transfer in R-1 and R-L. 
Transferring from any of the constrained, query-focused tasks outperforms transferring from unconstrained news summarization. 
Furthermore, transferring from WikiSum outperforms transfer from other datasets, which aligns with other work that shows the generalizability of Wikipedia as a source of data for task transfer \cite{fabbri-etal-2021-improving}.

\begin{table}[t!]
\centering
\small
\begin{tabular}{lccc}
\hline \textbf{Model} & \textbf{R-1} & \textbf{R-2} & \textbf{R-L} \\ \hline
Baselines & & \\
\hspace{3mm}\dyle~& \textit{34.42} & \textit{9.71} & \textit{30.10} \\
\hspace{3mm}\summn~& \textit{34.03} & \textit{9.28} & \textit{29.48} \\
\hspace{3mm}\bart & 31.87 & 9.08 & 27.50 \\
\hspace{3mm}\bart\wikisuffix & 32.68 & 8.97 & 28.74 \\
\hspace{3mm}\bart\wikisuffix~(Gold) & 39.54 & 15.65 & 35.17 \\
\hline
Two-stage & & \\
\hspace{3mm}\dpr  & 32.28 & 9.73 & 28.34 \\
\hspace{3mm}\twotower  & 33.02 & 10.17 & 28.90 \\
\hspace{3mm}\marge  & 31.99 & 8.97 & 27.93 \\
\hspace{3mm}\relregmodel  & 34.91 & 11.91 & 30.73 \\
\hspace{3mm}\relregmodel\wikisuffix & \textbf{36.45} & \textbf{12.81} & \textbf{32.28} \\
\hline
End-to-end & & \\
\hspace{3mm}\led & 34.18 & 10.32 & 29.95 \\
\hspace{3mm}\multiencmodel & 37.05 & 13.03 & 32.62 \\
\hspace{3mm}\multiencmodel\wikisuffix & \textbf{37.80} & \textbf{13.43} & \textbf{33.38} \\
\hline
\end{tabular}
\caption{ 
QMSum test-set performance of two-stage and end-to-end models that performed best on the validation set (Tables~\ref{tab:extractor_summ_val} and~\ref{tab:end2end}), including versions fine-tuned from the WikiSum-finetuned checkpoint (denoted by \textit{\mbox{-W}}).
%
Results reported in prior work are \textit{italicized}.
Also included is an extractive-oracle model that takes the gold spans (\S\ref{ssec:data}) as input. 
}
\label{tab:allmodels}
\end{table}

\subsection{Final Results}\label{ssec:final-results}
We now measure the test set performance of the best-performing architectures from \S\ref{ssec:two-stage-results} and \S\ref{ssec:end-to-end-results} in combination with the optimal transfer-learning approach from \S\ref{ssec:domain-adaptation}.
Results are presented in Table~\ref{tab:allmodels} along with baseline models.

We find that \relregmodel~and \multiencmodel~outperform existing state-of-the-art models by a substantial margin, and that initializing the model from the Wikisum-fine-tuned checkpoint further improves performance, with the best model exceeding current state-of-the-art performance by a difference of 3.38 R-1, 3.72 R-2, and 3.28 R-L. 
Comparing the best models from each category, we find that the end-to-end approach outperforms the two-stage.
Within the two-stage dual-encoder models, \twotower~ outperforms \dpr~on the test set despite the slightly worse performance on the validation set.  
We attribute this variation to the small size of the validation set, and our other findings remain consistent across validation and test sets. 
The single-encoder \relregmodel~outperforms the best dual-encoder model, with \relregmodel\wikisuffix~improving upon the current state-of-the-art performance by a difference of 2.03 R-1, 3.10 R-2, and 2.18 R-L. 
\section{Further Analysis}\label{sec:further-analysis}
In this section we conduct further analysis of the best performing models from Section~\ref{sec:results}.
First, we offer additional insights into the performance of those models on the QMSum dataset through a human-based study.
Next, we discuss the generalization abilities of those models by running experiments on the AQuaMuSe dataset.

\subsection{Human Evaluation}
\begin{table}[t!]
\centering
\small
\begin{tabular}{lcccc}
\hline \textbf{Model} & \textbf{Flu.} & \textbf{Rel.} & \textbf{Comp.} & \textbf{Fact.} \\ \hline
\bart~ & \textbf{4.08} & 3.68 & 3.22 & 3.31 \\ 
\relregmodel\wikisuffix~ &  3.87 & 3.81 & 3.67 & \textbf{3.70} \\ 
\multiencmodel\wikisuffix~ & 3.93 & \textbf{3.87} & \textbf{3.81} & 3.63 \\  \hline
\end{tabular}
\caption{Human evaluation of two best-performing models from Section \ref{sec:results}, along with a baseline \bart~model.
Summaries were evaluates across four dimensions: fluency (\textbf{Flu.}), relevance (\textbf{Rel.}), completeness (\textbf{Comp.}), and factuality (\textbf{Fact.}).
}
\label{tab:mturk-results}
\end{table}

To gain a better understanding of the performance of the models on the QMSum dataset, human judges were hired and asked to assess the quality of generated summaries.
Summaries were evaluated across four dimensions: 1) \textit{fluency}, measuring their grammatical quality, 2) \textit{relevance}, assessing their relevance to the input query, 3) \textit{completeness}, evaluating their comprehensiveness considering the input conversation and query, and 4) \textit{factuality}, measuring their factual consistency with respect to the conversation.
Scores were assigned on a Likert scale from 1 to 5 (best), where each example was evaluated by 3 judges with the final score averaged.
Results are presented in Table~\ref{tab:mturk-results}.

We find that the~\relregmodel\wikisuffix~and~\multiencmodel\wikisuffix~models achieved comparable performance across all of the evaluated dimensions, with summaries generated by~\multiencmodel\wikisuffix~rated as slightly more complete. 
The~\bart~baseline was rated highest in the fluency dimension, however, it was substantially outperformed by both of the introduced models on completeness and factuality.
One possible explanation for the slightly lower fluency scores for the~\relregmodel\wikisuffix~and~\multiencmodel\wikisuffix~models is that they
are better able to retrieve content from the source, which itself may have low fluency due to its conversational nature.
The results also highlight a gap between the performance of existing models and perfect scores, which shows that there is potential for improvement in future work.

\subsection{Dataset Generalization} 
\label{ssec:aquamuse-results}
\begin{table}[t!]
\centering
\small
\begin{tabular}{lccc}
\hline \textbf{Model} & \textbf{R-1} & \textbf{R-2} & \textbf{R-L} \\ \hline
\textit{Hi-MAP} & \textit{30.34} & \textit{14.82} & \textit{26.86} \\ 
\bart~ & 48.74 & 33.96 & 46.02 \\ 
\relregmodel\wikisuffix~ &  54.06 & 38.51 & 51.07 \\ 
\multiencmodel\wikisuffix~ & \textbf{63.62} & \textbf{51.27} & \textbf{61.37}  \\  \hline
\end{tabular}
\caption{AQuaMuSe test-set performance of two best-performing models from  \S\ref{sec:results}, along with a baseline \bart~model and previously reported results (in italics) for Hi-MAP  \cite{fabbri2019multinews}  from \citet{kulkarni2020aquamuse}. Note that the version of the dataset used for previous results would have been slightly different due to variations in document selection parameters and Common Crawl indices (see Appendix).
%
%
}

\label{tab:aquamuse-results}
\end{table}
To test that the automated evaluation results generalize beyond the QMSum dataset, we trained and evaluated the best-performing models on AQuaMuSe, another high-quality dataset for QFS that includes long documents (\S\ref{ssec:related-qfs}, \S\ref{ssec:data}). 
Test-set performance for the best-performing two-stage and end-to-end models, along with a baseline \bart~ model, are shown in Table~\ref{tab:aquamuse-results}.
Results are consistent with those for the QMSum dataset (Table~\ref{tab:allmodels}), with the best end-to-end model (\multiencmodel\wikisuffix) outperforming the best two-stage model (\relregmodel\wikisuffix), and both outperforming the baseline (\bart) model.
\section{Conclusion}
In this work, we conducted an exploratory study of neural models for query-focused summarization.
We studied two categories of models: two-stage and end-to-end, and presented two architectures, \relregmodel~and \multiencmodel, both of which improve ROUGE performance over prior state of the art by a substantial margin.
We also explored task-specific transfer learning, which further improved model performance.
Besides model performance, we discussed issues of computational efficiency that practitioners may factor into their modeling choices.
Finally, we conducted a human study suggesting that the summaries produced by the best-performing models are more factually correct and complete than a baseline model by a substantial margin.
We hope that the analysis and modeling contributions of this paper will be a resource for future research on query-focused summarization.
\section{Ethical Considerations}

\paragraph{Dataset Biases}
QMSum and AQuaMuSe contain meeting transcripts and documents in English and thus mainly represent the culture of the English-speaking populace. 
Political or gender biases may also exist in the dataset, and models trained on these datasets may propagate these biases
Additionally, the pretrained BART model carries biases from the data it was pretrained on. 
We did not stress test these models for biases and request that the users be aware of these potential issues in applying the models presented. 

\paragraph{Crowdsourcing Protocols}
 Workers were compensated \$1 per example, calibrated to equal a  \$12/hour payrate. 
We use the following qualifications to recruit MTurk workers with good track records: 
HIT approval rate greater than or equal to 97\%, number of HITs approved greater than or equal to 10000, and located in one of the following English native-speaking countries: Australia, Canada, New Zealand, United Kingdom, United States.
\paragraph{Misuse Potential and Failure Mode}
When properly used, the summarization models described in this paper can be time-saving.
However, the current model outputs may be factually inconsistent with the input documents, and in such a case could contribute to misinformation on the internet. 
This issue is present among all current abstractive summarization models and is an area of active research.

\paragraph{Environmental Cost}
The experiments described in the paper primarily make use of A100 GPUs. 
We typically used a single GPU per experiment, and the experiments may take up to a day when repeating across random seeds.
The largest backbone model used, \texttt{BART-Large}, has 400 million parameters. 
While our work required extensive experiments, future work and applications can draw upon our insights and need not repeat these comparisons. 

\section{Acknowledgements}
We thank Divyansh Agarwal for his insights on existing model architectures and assistance with training DPR. We thank Semih Yavuz for providing the initial implementation of the Fusion-in-Decoder model. 
\bibliography{anthology,custom}
\appendix
\section{Appendix}\label{sec:appendix}

\paragraph{Locator Model Parameters}
For \marge~ experiments, we apply the original fine-tuned BERT-base checkpoint from \citet{xu-lapata-2021-generating}, while for \dpr, we fine-tune a BERT-base model for both query and passage encoders following \citet{karpukhin-etal-2020-dense}. 

We report results for \relregmodel~ fine-tuned from an Electra-large checkpoint \cite{clark2020electra}. 
For a fair comparison with other metrics, we also fine-tuned \relregmodel~ from a BERT-base checkpoint. 
This version still outperformed \dpr~ by about a point in R-1, R-2, and R-L, demonstrating the advantage of this locator approach beyond the chosen base model. 

We apply \twotower~ fine-tuned from a distilled RoBERTa base \cite{liu2019roberta} checkpoint initially fine-tuned for the task of entailment. 
This approach of continuing fine-tuning from an entailment checkpoint is suggested by the sentence transformers library \cite{reimers-gurevych-2019-sentence}. 
We also experimented with fine-tuning the \twotower~ model from BERT-base and Electra-large checkpoints, but these locators did not perform better in initial experiments. 
\begin{table*}[t!]
\centering
\resizebox{\textwidth}{!}{%
\begin{tabular}{l|cccc|cccc|cccc|cccc||cccc|cccc|cccc|cccc||cc}
\hline
& \multicolumn{16}{c||}{\makecell{\textbf{Lexical Overlap b/w} \\ \textbf{Extractors and References}}} &
\multicolumn{16}{c||}{\makecell{\textbf{Lexical Overlap b/w} \\ \textbf{Extractors and Queries}}} &\multicolumn{2}{c}{\makecell{\textbf{Span Overlap b/w} \\ \textbf{Extractors and Golden Spans}}} \\ 
 \cline{2-35}
\textbf{Model} & \multicolumn{4}{c|}{\textbf{Top-1}} & \multicolumn{4}{c|}{\textbf{Top-5}} & \multicolumn{4}{c|}{\textbf{Top-15}} & \multicolumn{4}{c||}{\textbf{All}} & \multicolumn{4}{c|}{\textbf{Top-1}} & \multicolumn{4}{c|}{\textbf{Top-5}} & \multicolumn{4}{c|}{\textbf{Top-15}} & \multicolumn{4}{c||}{\textbf{All}} & \multicolumn{2}{c}{\textbf{All}} \\
 & \textit{R-1} & \textit{R-2} & \textit{R-L} & $\bar{x}$ & \textit{R-1} & \textit{R-2} & \textit{R-L} & $\bar{x}$ & \textit{R-1} & \textit{R-2} & \textit{R-L} & $\bar{x}$ & \textit{R-1} & \textit{R-2} & \textit{R-L} & $\bar{x}$ & \textit{R-1} & \textit{R-2} & \textit{R-L} & $\bar{x}$ & \textit{R-1} & \textit{R-2} & \textit{R-L} & $\bar{x}$ & \textit{R-1} & \textit{R-2} & \textit{R-L} & $\bar{x}$ & \textit{R-1} & \textit{R-2} & \textit{R-L} & $\bar{x}$ & \textit{Precision} & \textit{Recall} \\ \hline
\goldspans & 15.00 & 3.80 & 11.10 & 60 & \textbf{20.89} & \textbf{6.05} & \textbf{15.04} & 218 & \textbf{19.62} & \textbf{5.99} & \textbf{14.28} & 386 & \textbf{16.09} & \textbf{5.60} & \textbf{12.47} & 660 & 11.01 & 2.75 & 9.90 & 60 & 7.30 & 1.58 & 6.24 & 218 & 4.73 & 1.10 & 4.07 & 386 & 3.53 & 0.93 & 3.05 & 660 & 0.75 & 1.00
 \\
\lead & 8.17 & 0.98 & 6.30 & 82 & 12.84 & 1.69 & 9.17 & 309 & 13.13 & 1.81 & 9.21 & 463 & 8.77 & 1.79 & 6.77 & 978 & 4.88 & 0.60 & 4.49 & 82 & 5.51 & 0.72 & 4.71 & 309 & 3.76 & 0.64 & 3.26 & 463 & 1.70 & 0.37 & 1.55 & 978 & 0.09 & 0.20 \\ \hline
\dpr & 11.31 & 1.99 & 8.72 & 34 & 17.46 & 2.86 & 12.21 & 156 & 15.38 & 2.74 & 10.64 & 394 & 9.75 & 2.23 & 7.42 & 932 & 12.41 & 3.37 & 11.35 & 34 & 8.08 & 1.74 & 7.00 & 156 & 4.44 & 0.92 & 3.90 & 394 & 1.97 & 0.50 & 1.82 & 932 & 0.22 & 0.27 \\
\twotower & 23.67 & 3.34 & 15.66 & 82 & 16.13 & 3.35 & 11.18 & 413 & 9.65 & 2.58 & 7.31 & 930 & 9.16 & 2.52 & 6.99 & 994 & 9.63 & 1.58 & 8.26 & 82 & 3.49 & 0.83 & 3.09 & 413 & 1.81 & 0.50 & 1.65 & 930 & 1.66 & 0.46 & 1.53 & 994 & 0.07 & 0.24 \\ \hline
\marge & 7.13 & 0.72 & 5.81 & 20 & 13.76 & 1.39 & 10.22 & 92 & 14.85 & 1.74 & 11.09 & 269 & 9.21 & 1.52 & 7.16 & 896 & 7.22 & 0.81 & 6.88 & 20.61 & 6.86 & 0.67 & 6.09 & 92 & 4.70 & 0.61 & 4.20 & 269 & 1.84 & 0.36 & 1.70 & 896 & 0.15 & 0.21 \\
\relregmodel & \textbf{24.57} & \textbf{4.33} & \textbf{16.57} & 88 & 17.52 & 4.11 & 12.21 & 418 & 10.56 & 3.04 & 8.06 & 884 & 9.62 & 2.87 & 7.47 & 989 & 12.38 & 3.00 & 10.61 & 88 & 4.32 & 1.18 & 3.77 & 418 & 2.09 & 0.61 & 1.89 & 884 & 1.80 & 0.54 & 1.65 & 989 & 0.11 & 0.28 \\ 
\hline
\end{tabular}
}
\caption{
Performance of extractor models on the validation set.
The left and middle sections present the lexical overlap between utterances retrieved by extractor models and the reference summaries and summary queries, accordingly.
Lexical overlap is evaluated by means of ROUGE-1 (\textit{R-1)}, ROUGE-2 (\textit{R-2}), and ROUGE-L (\textit{R-L}) metrics.
Segments of the section focus on the lexical overlap between the highest ranked 1 (Top-1), 5 (Top-5), 15 (Top-15) utterances, and all utterances truncated to a 1024 token limit (All).
The table also includes the average word counts of all extracted utterances, denoted as $\bar{x}$.
The right section shows the span overlap between the utterance spans retrieved by the extractor models and those collected from human annotators by the authors of QMSum.
The performance is evaluated by means of \textit{Precision} and \textit{Recall} scores and uses the highest ranked utterances truncated to the limit of 1024 tokens.
}
\label{tab:extractor_models_full}
\end{table*}

\paragraph{Summarization Model Parameters}
In all experiments described in this work, the LED model was initialized from the \texttt{allenai/led-large-16384} checkpoint.
Two model hyperparameters, \textit{maximal input size} and \textit{attention window size}, were chosen through a hyperparameter search with candidate models selected based on their performance on the validation set. 
Best hyperparamters were found to be: \texttt{16384} maximum input size, and \texttt{1024} attention window size. 
LED models were trained for \texttt{10} epochs, with a batch size \texttt{1}, gradient accumulation set to \texttt{4} steps, and learning rate set to \texttt{0.000005}.
The \multiencmodel~ model was initialized from the \texttt{facebook/bart-large} checkpoint. The model hyperparameters, \textit{maximal input size} and \textit{attention window size}, were chosen through a hyperparameter search with candidate models selected based on their performance on the validation set, with results reported in the paper.
Best hyperparamters were found to be: \texttt{16384} maximum input size, and \texttt{512} attention window size. 
The \multiencmodel~ models were trained for \texttt{10} epochs, with a batch size of \texttt{1} and learning rate set to \texttt{0.000005}.

\paragraph{QMSum Details}
QMSum contains 1,808 query-summary pairs in total, with a train/validation/test split of 1257/272/281.
It is made available through an MIT license\footnote{\url{https://github.com/Yale-LILY/QMSum/blob/main/LICENSE}}, which aligns with our use for research purposes.
Non-identifying names are used in place of real names.

\paragraph{AQuaMuse Details}
We experiment the V3, abstractive version of AQuaMuse, consisting of 7725 query-summary pairs, with a train/validation/test split of 5566/596/734.
The original AQuaMuse paper reported results on V2 of the dataset, which contains a slightly different input document set due to variations in the semantic overlap threshold used to retrieve documents.
Some input documents could not be retrieved due to differences in the Common Crawl index used; we use the cleaned, reproduced version of the C4 dataset \cite{2020t5} from the Common Crawl made available by AI2\footnote{\url{https://github.com/allenai/allennlp/discussions/5056}}.
We kept examples for which all input documents were found, which resulted in a dataset of 6896 examples. 
The natural language questions it contains are made available through an Apache 2.0 license\footnote{\url{https://github.com/google-research-datasets/natural-questions/blob/master/LICENSE}}, which aligns with our use for research purposes.
This dataset uses publicly available entities from Wikipedia.

\section{Human Annotation Interface}
\begin{figure*} [ht]
  \centering
  \includegraphics[width=0.95\textwidth]{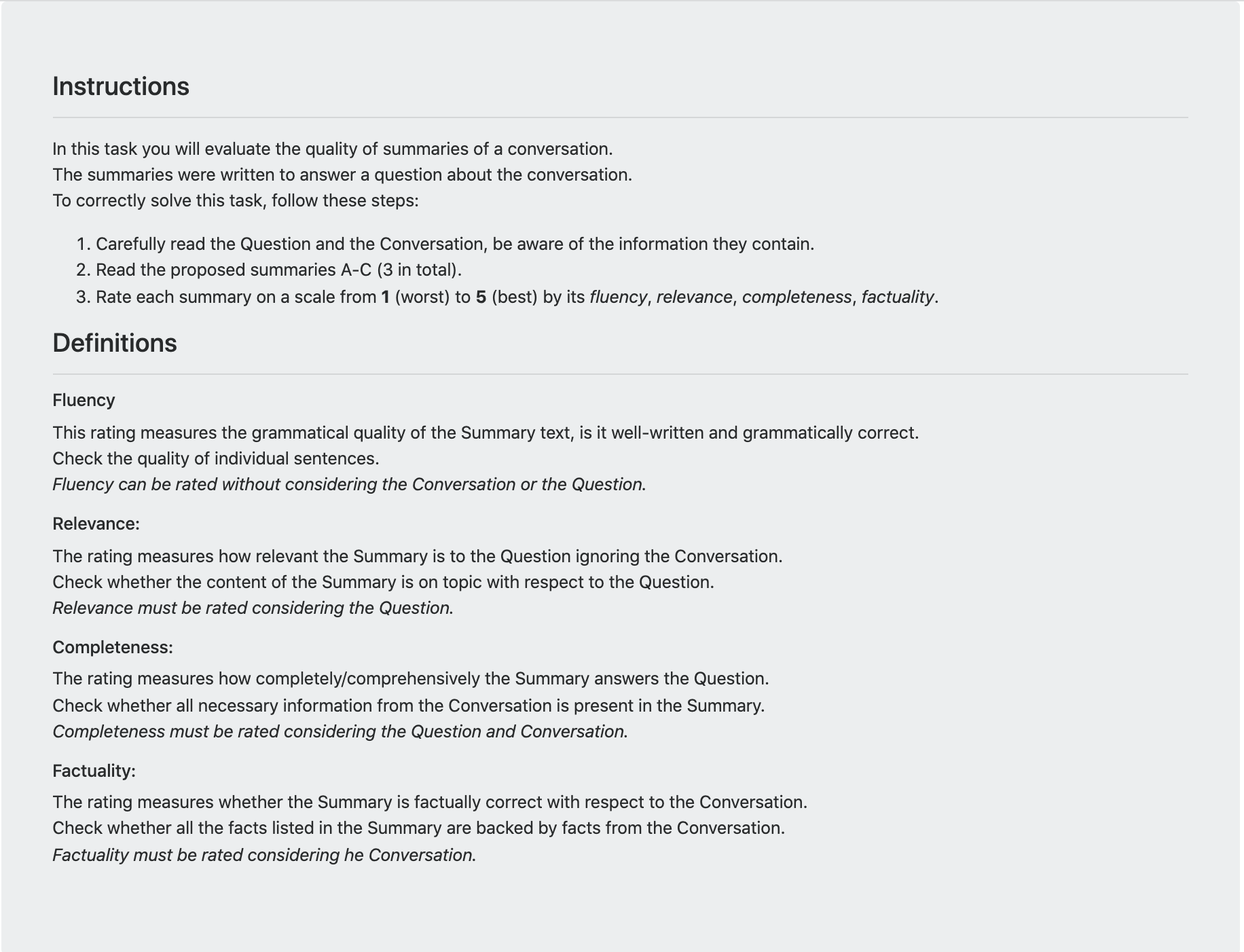}
    \caption{Instructions presented to annotators for the human studies}
  \label{fig:mturk-ui}
\end{figure*}

The instructions shown to the annotators during human studies are presented in Figure~\ref{fig:mturk-ui}

\end{document}